\documentclass[conference]{ieeeconf}
\IEEEoverridecommandlockouts

\usepackage[backend=bibtex,
            hyperref=false,
            url=false,
            isbn=false,
            doi=false,
            backref=false,
            style=ieee,
            citestyle=numeric-comp,
            sorting=nyt,
            block=none]{biblatex}

\renewcommand{\bibfont}{\small}
\usepackage{flushend}
\usepackage{balance}
\usepackage{graphics}
\usepackage[pdftex]{graphicx}
\usepackage{wrapfig}
\DeclareGraphicsExtensions{.pdf,.png,.jpg}

\usepackage[skip=2pt,font=small]{caption}
\usepackage{subcaption}
\usepackage[rightcaption]{sidecap}
\usepackage{pbox}

\usepackage{mathtools}
\usepackage{amsmath, amssymb, amscd}
\usepackage{ wasysym } 
\usepackage{amsfonts}
\usepackage{mathptmx} 
\usepackage{gensymb} 
\usepackage{nicefrac}       

\DeclareMathAlphabet{\mathcal}{OMS}{lmsy}{m}{n}
\DeclareSymbolFont{largesymbols}{OMX}{cmex}{m}{n}
\usepackage{textcomp} 

\usepackage{algorithm} 
\usepackage{algorithmicx, algpseudocode} 

\usepackage{array} 
\usepackage{tabularx}
\usepackage{multirow}
\usepackage{multicol}
\usepackage{booktabs}

\usepackage[T1]{fontenc} 
\usepackage[utf8]{inputenc}
\usepackage[english]{babel} 
\usepackage{units}
\usepackage{bm}
\usepackage{times} 
\usepackage{xspace}
\usepackage{flushend}
\usepackage{csquotes}
\usepackage{makeidx}
\usepackage{blindtext}
\let\labelindent\relax
\usepackage{enumitem}

\usepackage{soul} 
\usepackage{subfiles} 

\usepackage[yyyymmdd]{datetime}

\date{\protect\formatdate{1}{1}{2001}}

\usepackage{url}
\makeatletter
\g@addto@macro{\UrlBreaks}{\UrlOrds}
\makeatother
\usepackage{color}
\usepackage[usenames,dvipsnames,table]{xcolor}

\usepackage{marginnote}
\usepackage{soul} 

 \usepackage[disable]{todonotes}
\newcommand{\tocite}[1]{%
\textcolor{red}{[cite:\ifthenelse{\equal{#1}{}}{}{#1}?]}
}

\newcommand{\ignore}[1]{}

\newcommand{\figref}[1]{Fig.    \,\ref{fig:#1}}

\newcommand{\secref}[1]{Section\,\ref{sec:#1}}

\newcommand{\piparams}{\mathbf{\theta}_\pi}
\newcommand{\repparams}{\mathbf{\theta}_s}

\newcommand{\markov}{\mathcal{M}}

\newcommand{\tstate}{\mathbf{s}_t}
\newcommand{\states}{\mathcal{S}}
\newcommand{\actions}{\mathcal{A}}

\newcommand{\taction}{\mathbf{a}_t}
\newcommand{\stateinitial}{\rho_0}
\newcommand{\rewardfn}{r}
\newcommand{\transition}{\mathcal{T}}
\newcommand{\timemax}{T}
\newcommand{\policy}{\pi}

\newcommand{\cartdelta}{\Delta \mathbf{x}}
\newcommand{\cartpos}{\mathbf{x}}
\newcommand{\cartpost}{\mathbf{x}_t}
\newcommand{\cartposd}{\mathbf{x}_{\textrm{des}}}
\newcommand{\cartposdt}{\mathbf{x}_k}
\newcommand{\cartvel}{\mathbf{v}}
\newcommand{\cartveld}{\mathbf{v}_{\textrm{des}}}
\newcommand{\cartveldt}{\mathbf{v}_k}

\newcommand{\cartaccd}{\mathbf{a}_{\textrm{des}}}
\newcommand{\cartaccdt}{\mathbf{a}_k}
\newcommand{\cartaccu}{\mathbf{a}_u}

\newcommand{\kp}{\mathbf{k}_{\textrm{p}}}
\newcommand{\kv}{\mathbf{k}_{\textrm{v}}}
\newcommand{\taurobot}{\mathbf{\tau}_{\textrm{u}}}

\makeatletter
\newenvironment{mcases}[1][l]
 {\let\@ifnextchar\new@ifnextchar
  \left\lbrace
  \array{@{}l@{\quad}#1@{}}}
 {\endarray\right.}
\makeatother

\usepackage{algorithm}
\usepackage{algpseudocode}

\title{\LARGE {\bf
Making Sense of Vision and Touch}: Self-Supervised Learning of Multimodal Representations for Contact-Rich Tasks
}

\author{%
Michelle A. Lee$^{*}$,
Yuke Zhu$^{*}$,
Krishnan Srinivasan,
Parth Shah,
\\Silvio Savarese,
Li Fei-Fei,
Animesh Garg,
Jeannette Bohg
\thanks{\vspace{-10pt} \hrule \vspace{1pt} $^{*}$Authors have contributed equally and names are in alphabetical order.}%
\thanks{Authors are with the Department of Computer Science, Stanford University. {\tt\scriptsize [mishlee,yukez,krshna,pshah9,ssilvio,feifeili, animeshg,bohg]@stanford.edu}. A. Garg is also at Nvidia, USA.}
\thanks{This work has been partially supported by JD.com American Technologies Corporation (“JD”) under the SAIL-JD AI Research Initiative and by the Toyota Research Institute ("TRI"). This article solely reflects the opinions and conclusions of its authors and not of JD, any entity associated with JD.com, TRI, or any entity associated with Toyota. We are grateful to Oussama Khatib for lending the Kuka IIWA, as well as to Shameek Ganguly and Mikael Jorda for insightful research discussions.}%
}

\begin{document}
\maketitle
\thispagestyle{empty}
\pagestyle{empty}

\begin{abstract}
Contact-rich manipulation tasks in unstructured environments often require both haptic and visual feedback. However, it is non-trivial to manually design a robot controller that combines modalities with very different characteristics. While deep reinforcement learning has shown success in learning control policies for high-dimensional inputs, these algorithms are generally intractable to deploy on real robots due to sample complexity. We use self-supervision to learn a compact and multimodal representation of our sensory inputs, which can then be used to improve the sample efficiency of our policy learning. We evaluate our method on a peg insertion task, generalizing over different geometry, configurations, and clearances, while being robust to external perturbations. We present results in simulation and on a real robot.  
\end{abstract}

\section{Introduction}

Even in routine tasks such as inserting a car key into the ignition, humans effortlessly combine the senses of vision and touch to complete the task. Visual feedback provides semantic and geometric object properties for accurate reaching or grasp pre-shaping. Haptic feedback provides observations of current contact conditions between object and environment for accurate localization and control under occlusions. These two feedback modalities are complementary and concurrent during contact-rich manipulation~\cite{blake2004neural}. Yet, there are few algorithms that endow robots with a similar ability. While the utility of multimodal data has frequently been shown in robotics~\cite{bicchi1988integrated,romano2011human,veiga2015stabilizing,Song:2014}, the proposed manipulation strategies are often task-specific. While learning-based methods do not require manual task specification, the majority of learned manipulation policies close the control loop around a single modality, often vision \cite{Levine:Finn:2016,chebotar2017path, finn2017deep, zhu2018reinforcement}.

\begin{figure}[t!]
\centering
\includegraphics[width=\columnwidth]{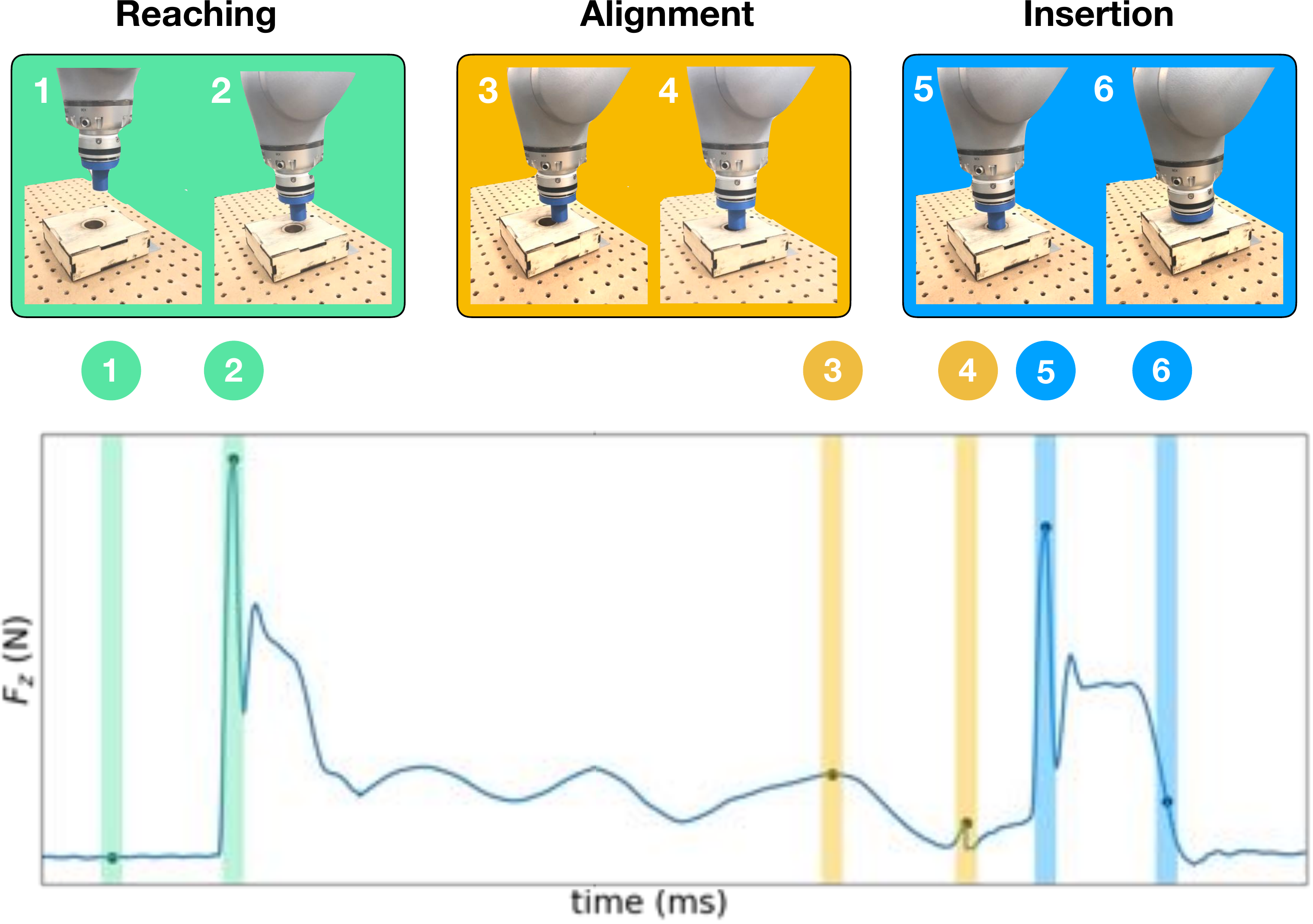}
\caption{Force sensor readings in the z-axis (height) and visual observations are shown with corresponding stages of a peg insertion task. The force reading transitions from (1) the arm moving in free space to (2) making contact with the box. While aligning the peg, the forces capture the sliding contact dynamics on the box surface (3, 4). Finally, in the insertion stage, the forces peak as the robot attempts to insert the peg at the edge of the hole (5), and decrease when the peg slides into the hole (6).}
\label{fig:insertion}
\end{figure}

In this work, we equip a robot with a policy that leverages multimodal feedback from vision and touch, two modalities with very different dimensions, frequencies, and characteristics. This policy is learned through self-supervision and generalizes over variations of the same contact-rich manipulation task in geometry, configurations, and clearances. It is also robust to external perturbations. Our approach starts with using neural networks to learn a shared representation of haptic and visual sensory data, as well as proprioceptive data. Using a self-supervised learning objective, this network is trained to predict optical flow, whether contact will be made in the next control cycle, and concurrency of visual and haptic data. The training is action-conditional to encourage the encoding of action-related information. The resulting compact representation of the high-dimensional and heterogeneous data is the input to a policy for contact-rich manipulation tasks using deep reinforcement learning.
The proposed decoupling of state estimation and control achieves practical sample efficiency for learning both representation and policy on a real robot. 
Our primary contributions are:
\begin{enumerate}[
    topsep=0pt,
    noitemsep,
    leftmargin=*,
    itemindent=12pt]
\item A model for multimodal representation learning from which a contact-rich manipulation policy can be learned. 
\item Demonstration of insertion tasks that effectively utilize both haptic and visual feedback for hole search, peg alignment, and insertion (see Fig \ref{fig:insertion}). Ablative studies compare the effects of each modality on task performance. 
\item Evaluation of generalization to tasks with different peg geometry and of robustness to perturbation and sensor noise. 
\end{enumerate}

\section{Related Work and Background}
\subsection{Contact-Rich Manipulation}
Contact-rich tasks, such as peg insertion, block packing, and edge following, have been studied for decades due to their relevance in manufacturing. Manipulation policies often rely entirely on haptic feedback and force control, and assume sufficiently accurate state estimation~\cite{Whitney:1987}. They typically generalize over certain task variations, for instance, peg-in-chamfered-hole insertion policies that work independently of peg diameter~\cite{Whitney1982}. However, entirely new policies are required for new geometries. For chamferless holes, manually defining a small set of viable contact configurations has been successful~\cite{Caine99} but cannot accommodate the vast range of real-world variations. \cite{Song:2014} combines visual and haptic data for inserting two planar pegs with more complex cross sections, but assumes known peg geometry.

Reinforcement learning approaches have recently been proposed to address variations in geometry and configuration for manipulation.
\cite{Levine:Finn:2016,zhu2018reinforcement} trained neural network policies using RGB images and proprioceptive feedback. Their approach works well in a wide range of tasks, but the large object clearances compared to automation tasks may explain the sufficiency of RGB data. A series of learning-based approaches have relied on haptic feedback for manipulation. Many of them are concerned with estimating the stability of a grasp before lifting an object~\cite{calandra2017feeling,Yasemin:2013}, even suggesting a regrasp~\cite{Yevgen:Karol:2015}. Only a few approaches learn entire manipulation policies through reinforcement only given haptic feedback~\cite{Mrinal:2011, sung2017learning,van2016stable, van2015learning}. While \cite{Mrinal:2011} relies on raw force-torque feedback, \cite{sung2017learning,van2016stable} learn a low-dimensional representation of high-dimensional tactile data before learning a policy. Even fewer approaches exploit the complementary nature of vision and touch. Some of them extend their previous work on grasp stability estimation~\cite{YaseminRenaud,Calandra:2018}. Others perform full manipulation tasks based on multiple input modalities~\cite{Kappler-RSS-15} but require a pre-specified manipulation graph and demonstrate only on a single task, or require human demonstration and object CAD models~\cite{abu2015adaptation}. There have been promising works that train manipulation policy in simulation and transfer them to a real robot~\cite{andrychowicz2018learning, peng2018sim, bousmalis2018using}. However, only few works focused on contact-rich tasks~\cite{fu2016one} and none relied on haptic feedback in simulation, most likely because of the lack of fidelity of contact simulation and collision modeling for articulated rigid-body systems~\cite{shameekcollisions,fazeli2017fundamental}. 

\subsection{Multimodal Representation Learning}\label{sec:related:repr}
The complementary nature of heterogeneous sensor modalities has previously been explored for inference and decision making. The diverse set of modalities includes vision, range, audio, haptic and proprioceptive data as well as language. This heterogeneous data makes the application of hand-designed features and sensor fusion extremely challenging. That is why learning-based methods have been on the forefront. 
\cite{Calandra:2018,gao2016deep,YaseminRenaud,SinapovSS14} are examples of fusing visual and haptic data for grasp stability assessment, manipulation, material recognition, or object categorization. \cite{liu2017learning,sung2017learning} fuse vision and range sensing and \cite{sung2017learning} adds language labels. While many of these multimodal approaches are trained through a classification objective~\cite{Calandra:2018,gao2016deep,YaseminRenaud,yang2017deep}, in this paper we are interested in multimodal representation learning for control. 
A popular representation learning objective is reconstruction of the raw sensory input~\cite{de2018integrating,StateReprLearning,van2016stable, yang2017deep}.  This unsupervised objective benefits learning stability and speed, but it is also data intensive and prone to overfitting~\cite{de2018integrating}. When learning for control, action-conditional predictive representations can encourage the state representations to capture action-relevant information~\cite{StateReprLearning}. Studies attempted to predict full images when pushing objects with benign success~\cite{agrawal2016learning, babaeizadeh2017stochastic, oh2015action}. In these cases either the underlying dynamics is deterministic~\cite{oh2015action}, or the control runs at a low frequency~\cite{finn2017deep}. In contrast, we operate with haptic feedback at 1\unit{kHz} and send Cartesian control commands at 20\unit{Hz}. We use an action-conditional surrogate objective for predicting optical flow and contact events with self-supervision.


There is compelling evidence that the interdependence and concurrency of different sensory streams aid perception and manipulation \cite{edelman1987neural,lacey2016crossmodal,2016_TRO_IP}. However, few studies have explicitly exploited this concurrency in representation learning. Examples include \cite{srivastava2012multimodal} for visual prediction tasks and \cite{ngiam2011multimodal,owens2018audio} for audio-visual coupling. Following \cite{owens2018audio}, we propose a self-supervised objective to fuse visual and haptic data.
\section{Problem Statement and Method Overview}
\label{sec:ps}
Our goal is to learn a policy on a robot for performing contact-rich manipulation tasks. We want to evaluate the value of combining multisensory information and the ability to transfer multimodal representations across tasks. For sample efficiency, we first learn a neural network-based feature representation of the multisensory data. The resulting compact feature vector serves as input to a policy that is learned through reinforcement learning.

We model the manipulation task as a finite-horizon, discounted Markov Decision Process (MDP) $\markov$, with a state space $\states$, an action space $\actions$, state transition dynamics $\transition : \states \times \actions \to \states$, an initial state distribution $\stateinitial$, a reward function $\rewardfn : \states \times \actions \to \mathbb{R}$, horizon $\timemax$, and discount factor $\gamma \in (0, 1]$.
To determine the optimal stochastic policy $\policy : \states \to \mathbb{P}(\actions)$, we want to maximize the expected discounted reward 
\begin{equation}
J(\policy) = \mathbb{E}_\policy \left[\sum^{\timemax-1}_{t=0} \gamma \rewardfn(\tstate, \taction) \right]
\label{eqn:loss}
\end{equation}
We represent the policy by a neural network with parameters $\piparams$ that are learned as described in Sec.~\ref{sec:policy-control}. $\states$ is defined by the low-dimensional representation learned from high-dimensional visual and haptic sensory data. This representation is a neural network parameterized by  $\repparams$ and is trained as described in Sec.~\ref{sec:representation-learning}. $\actions$ is defined over continuously-valued, 3D displacements $\cartdelta$ in Cartesian space. The controller design is detailed in Sec.~\ref{sec:policy-control}. 

\begin{figure*}[t!]
\centering
\includegraphics[width=\linewidth]{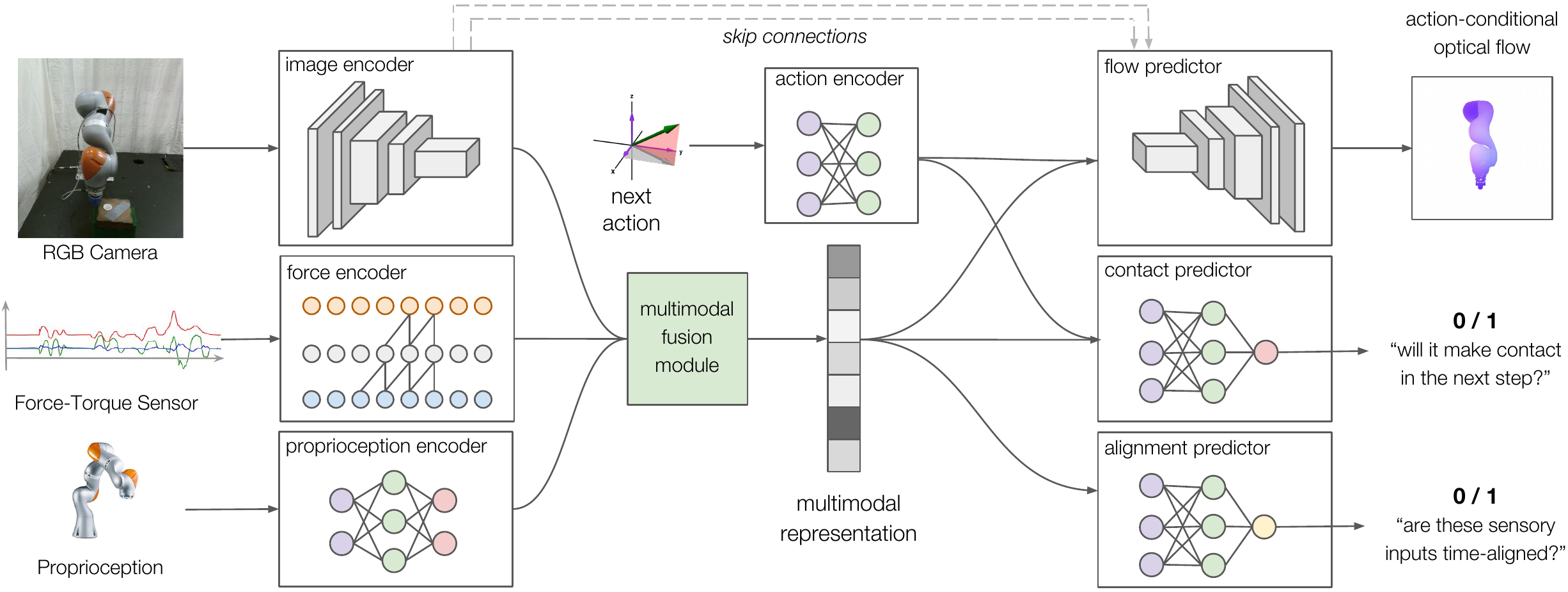}
\caption{Neural network architecture for multimodal representation learning with self-supervision. The network takes data from three different sensors as input: RGB images, F/T readings over a 32\unit{ms} window, and end-effector position and velocity. It encodes and fuses this data into a multimodal representation based on which controllers for contact-rich manipulation can be learned. This representation learning network is trained end-to-end through self-supervision.}
\label{fig:network_architecture}
\end{figure*}

\section{Multi-Modal Representation Model}
\label{sec:representation-learning}
Deep networks are powerful tools to learn representations from high-dimensional data~\cite{lecun2015deep} but require a substantial amount of training data. Here, we address the challenge of seeking sources of supervision that do not rely on laborious human annotation. We design a set of predictive tasks that are suitable for learning visual and haptic representations for contact-rich manipulation tasks, where supervision can be obtained via automatic procedures rather than manual labeling. \figref{network_architecture} visualizes our representation learning model.

\subsection{Modality Encoders}
Our model encodes three types of sensory data available to the robot: RGB images from a fixed camera, haptic feedback from a wrist-mounted force-torque (F/T) sensor, and proprioceptive data from the joint encoders of the robot arm. The heterogeneous nature of this data requires domain-specific encoders to capture the unique characteristics of each modality. For visual feedback, we use a 6-layer {\em convolutional neural network\/} (CNN) similar to FlowNet~\cite{flownet1} to encode $128\times 128\times 3$ RGB images. We add a fully-connected layer to transform the final activation maps into a $128$-d feature vector. For haptic feedback, we take the last 32 readings from the six-axis F/T sensor as a $32\times 6$ time series and perform 5-layer causal convolutions~\cite{oord2016wavenet} with stride 2 to transform the force readings into a $64$-d feature vector. For proprioception, we encode the current position and velocity of the end-effector with a 2-layer {\em multilayer perceptron\/} (MLP) to produce a $32$-d feature vector. The resulting three feature vectors are concatenated into one vector and passed through the multimodal fusion module (2-layer MLP) to produce the final $128$-d multimodal representation.

\subsection{Self-Supervised Predictions} 
The modality encoders have nearly half a million learnable parameters and require a large amount of labeled training data. To avoid manual annotation, we design training objectives for which labels can be automatically generated through self-supervision. Furthermore, representations for control should encode the action-related information. To achieve this, we design two action-conditional representation learning objectives. Given the next robot action and the compact representation of the current sensory data, the model has to predict (i) the optical flow generated by the action and (ii) whether the end-effector will make contact with the environment in the next control cycle. Ground-truth optical flow annotations are automatically generated given proprioception and known robot kinematics and geometry~\cite{flownet1,GarciaCifuentes.RAL}. Ground-truth annotations of binary contact states are generated by applying simple heuristics on the F/T readings. 

The next action, i.e. the end-effector motion, is encoded by a 2-layer MLP. Together with the multimodal representation it forms the input to the flow and contact predictor. The flow predictor uses a 6-layer convolutional decoder with upsampling to produce a flow map of size $128\times 128\times 2$. Following~\cite{flownet1}, we use 4 skip connections. The contact predictor is a 2-layer MLP and performs binary classification.

As discussed in Sec.~\ref{sec:related:repr}, there is concurrency between the different sensory streams leading to correlations and redundancy, e.g., seeing the peg, touching the box, and feeling the force. We exploit this by introducing a third representation learning objective that predicts whether two sensor streams are temporally aligned~\cite{owens2018audio}. During training, we sample a mix of time-aligned multimodal data and randomly shifted ones. The alignment predictor (a 2-layer MLP) takes the low-dimensional representation as input and performs binary classification of whether the input was aligned or not.

We train the action-conditional optical flow with endpoint error (EPE) loss averaged over all pixels~\cite{flownet1}, and both the contact prediction and the alignment prediction with cross-entropy loss. During training, we minimize a sum of the three losses end-to-end with stochastic gradient descent on a dataset of rolled-out trajectories. Once trained, this network produces a $128$-d feature vector that compactly represents multimodal data. This vector from the input to the manipulation policy learned via reinforcement learning.

\begin{figure}[t!]
\centering
\includegraphics[width=\linewidth,clip]{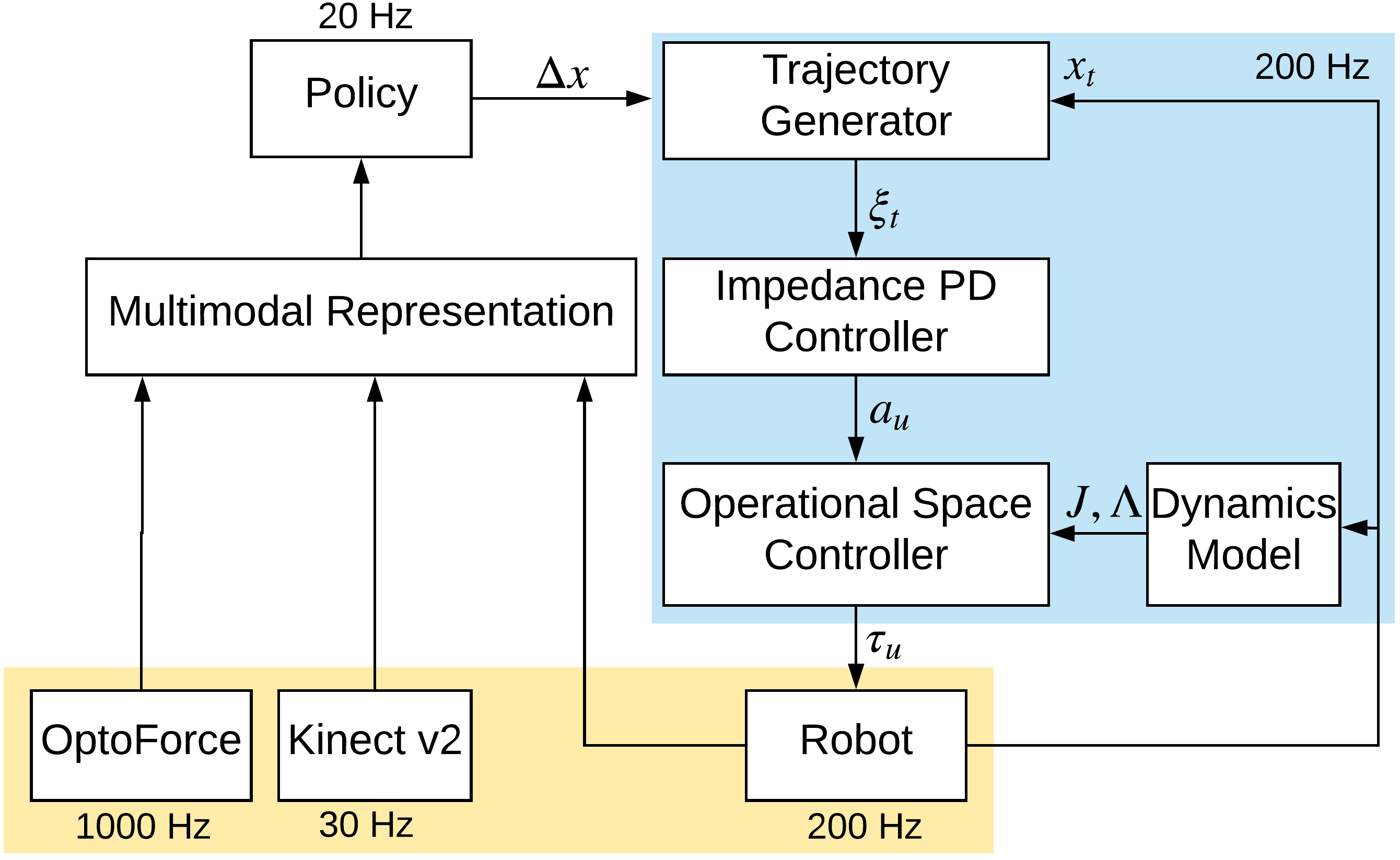}
\caption{Our controller takes end-effector position displacements from the policy at 20\unit{Hz} and outputs robot torque commands at 200\unit{Hz}. The trajectory generator interpolates high-bandwidth robot trajectories from low-bandwidth policy actions. The impedance PD controller tracks the interpolated trajectory. The operational space controller uses the robot dynamics model to transform Cartesian-space accelerations into commanded joint torques. The resulting controller is compliant and reactive.}
\label{fig:controller}
\end{figure}

\section{Policy Learning and Controller Design}
\label{sec:policy-control}
Our final goal is to equip a robot with a policy for performing contact-rich manipulation tasks that leverage multimodal feedback. Though it is possible to engineer controllers for specific instances of these tasks~\cite{Whitney:1987,Song:2014}, this effort is difficult to scale to the large variability of real-world tasks. Therefore, it is desirable to enable a robot to supervise itself where the learning process is applicable to a broad range of tasks. Given its recent success in continuous control~\cite{lillicrap2015continuous,schulman2015trust}, deep reinforcement learning lends itself well to learning policies that map high-dimensional features to control commands. 

\noindent \textbf{Policy Learning.}
Modeling contact interactions and multi-contact planning still result in complex optimization problems~\cite{Posa:2013ez,ponton2016,tonneau18} that remain sensitive to inaccurate actuation and state estimation. We formulate contact-rich manipulation as a model-free reinforcement learning problem to investigate its performance when relying on multimodal feedback and when acting under uncertainty in geometry, clearance and configuration. By choosing model-free, we also eliminate the need for an accurate dynamics model, which is typically difficult to obtain in the presence of rich contacts. Specifically, we choose trust-region policy optimization (TRPO)~\cite{schulman2015trust}. TRPO imposes a bound of KL-divergence for each policy update by solving a constrained optimization problem, which prevents the policy from moving too far away from the previous step. The policy network is a 2-layer MLP that takes as input the $128$-d multimodal representation and produces a 3D displacement $\cartdelta$ of the robot end-effector. To train the policy efficiently, we freeze the representation model parameters during policy learning, such that it reduces the number of learnable parameters to $3\%$ of the entire model and substantially improves the sample efficiency. 


\noindent \textbf{Controller Design.} Our controller takes as input Cartesian end-effector displacements $\cartdelta$ from the policy at 20\unit{Hz}, and outputs direct torque commands $\taurobot$ to the robot at 200\unit{Hz}. Its architecture can be split into three parts: trajectory generation, impedance control and operational space control (see Fig~\ref{fig:controller}). Our policy outputs Cartesian control commands instead of joint-space commands, so it does not need to implicitly learn the non-linear and redundant mapping between 7-DoF joint space and 3-DoF Cartesian space. We use direct torque control as it gives our robot compliance during contact, which makes the robot safer to itself, its environment, and any nearby human operator. In addition, compliance makes the peg insertion task easier to accomplish under position uncertainty, as the robot can slide on the surface of the box while pushing downwards \cite{Mrinal:2011,Righetti2014,Eppner:2015:EEC:2879361.2879370}.

The trajectory generator bridges low-bandwidth output of the policy (limited by the forward pass of our representation model), and the high-bandwidth torque control of the robot. Given $\cartdelta$ from the policy and the current end-effector position $\cartpost$, we calculate the desired end-effector position  $\cartposd$. The trajectory generator interpolates between $\cartpost$ and $\cartposd$ to yield a trajectory $\xi_t = \{\cartposdt, \cartveldt, \cartaccdt \}_{k=t}^{t+T}$ of end-effector position, velocity and acceleration at 200\unit{Hz}. This forms the input to a PD impedance controller to compute a task space acceleration command:
$\cartaccu = \cartaccd -\kp (\cartpos-\cartposd) - \kv (\cartvel - \cartveld),$ where $\kp$ and $\kv$ are manually tuned gains.

By leveraging known kinematic and dynamics models of the robot, we can calculate joint torques from Cartesian space accelerations with the dynamically-consistent operational space formulation~\cite{Khatib1995a}. We compute the force at the end-effector with $\mathbf{F}= \Lambda \cartaccu$, where $\Lambda$ is the inertial matrix in the end-effector frame that decouples the end-effector motions. Finally, we map from $\mathbf{F}$ to joint torque commands with the end-effector Jacobian $J$, which is a function of joint angle $\mathbf{q}$: $\taurobot = J^T(\mathbf{q})\mathbf{F}$. 
\begin{figure*}[t!]
\vspace{-10pt}
\centering
\begin{subfigure}{.48\textwidth}
    \centering
    \includegraphics[width=.9\linewidth]{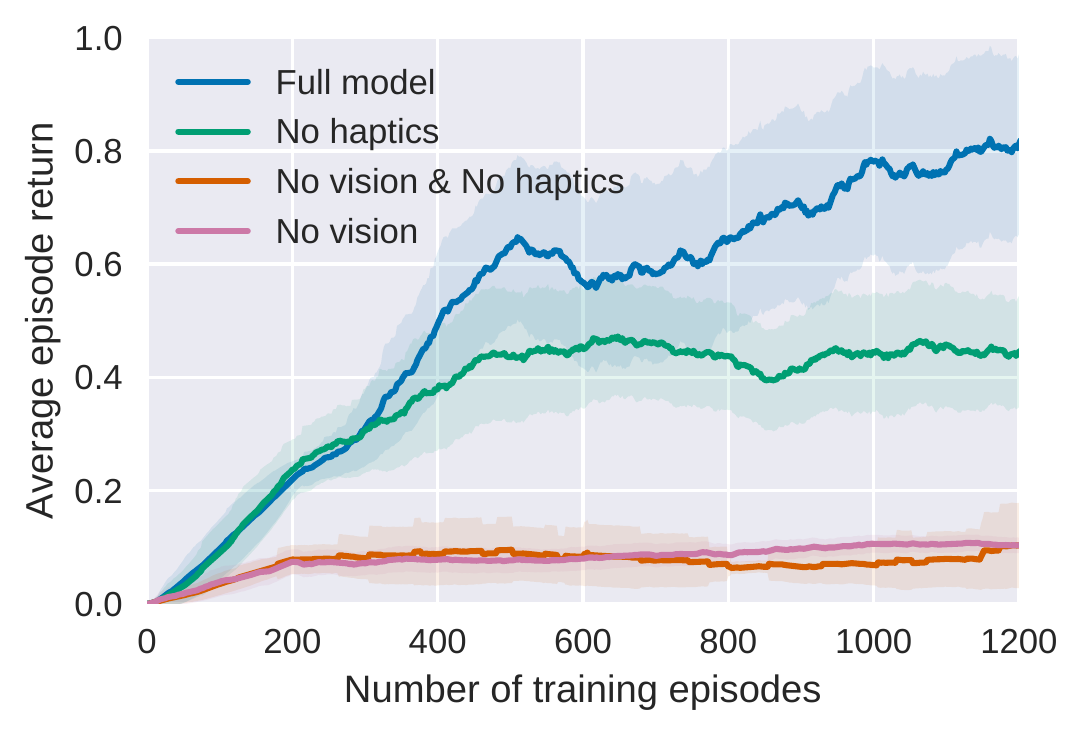}    
    \caption{Training curves of reinforcement learning}
    \label{fig:training_curve}
\end{subfigure}
\begin{subfigure}{.50\textwidth}
    \centering
    \includegraphics[width=.9\linewidth]{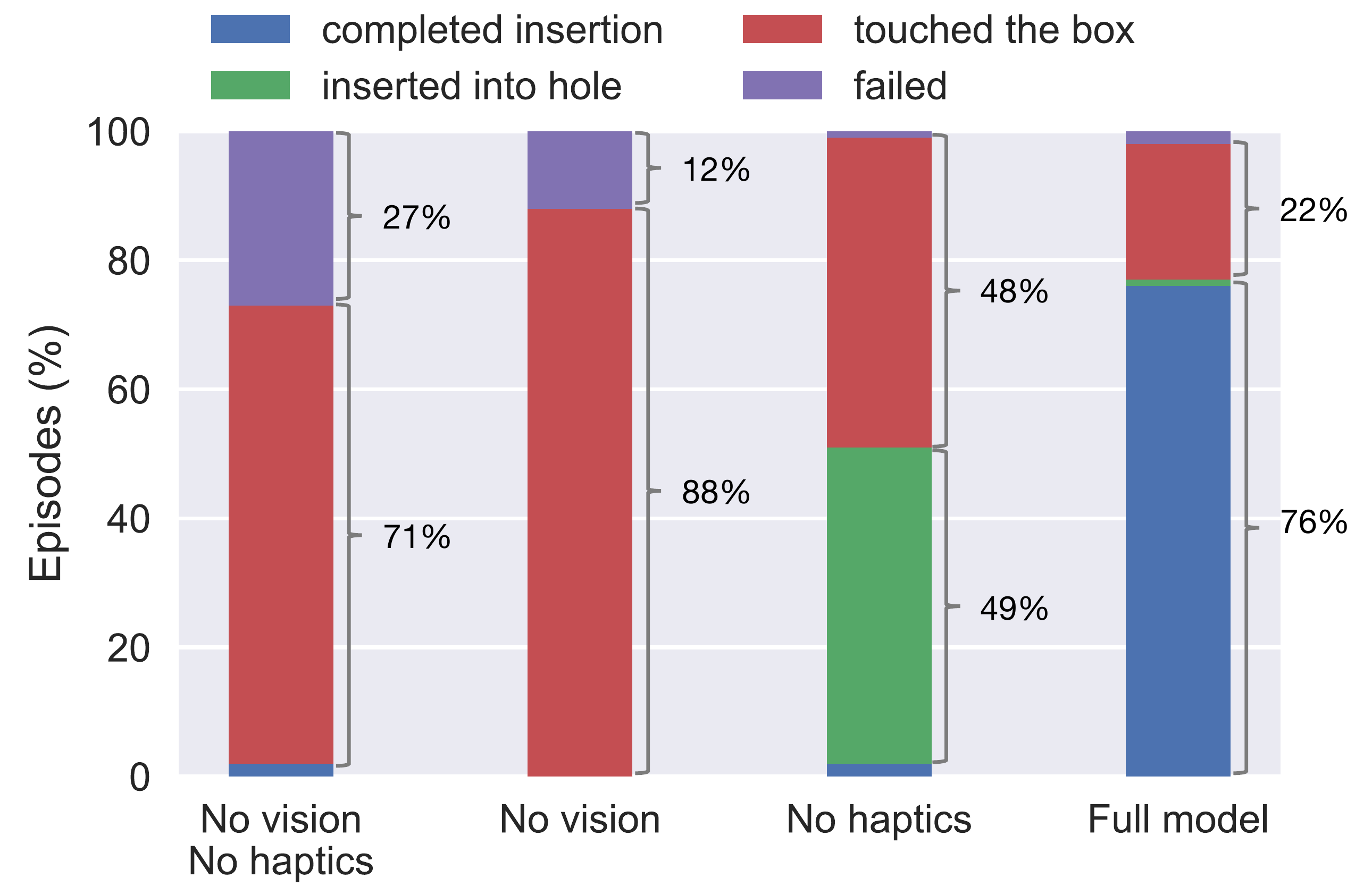}
    \caption{Policy evaluation statistics}
    \label{fig:policy_statistics}
\end{subfigure}
\caption{Simulated Peg Insertion: Ablative study of representations trained on different combinations of sensory modalities. We compare our full model, trained with a combination of visual and haptic feedback and proprioception, with baselines that are trained without vision, or haptics, or either.  (b) The graph shows partial task completion rates with different feedback modalities, and we note that both the visual and haptic modalities play an integral role for contact-rich tasks.}
\label{fig:quantitative_results}
\end{figure*}

\section{Experiments: Design and Setup}

The primary goal of our experiments is to examine the effectiveness of the multimodal representations in contact-rich manipulation tasks. 
In particular, we design the experiments to answer the following three questions: 1) What is the value of using {\em all\/} instead of a subset of modalities? 2) Is policy learning on the real robot \textit{practical} with a learned representation? 3) Does the learned representation \textit{generalize} over task variations and recover from perturbations?

\noindent \textbf{Task Setup.} 
We design a set of peg insertion tasks where task success requires joint reasoning over visual and haptic feedback. 
We use five different types of pegs and holes fabricated with a 3D printer: round peg, square peg, triangular peg, semicircular peg, and hexagonal peg, each with a nominal clearance of around 2mm as shown in \figref{peg_types}. 

\noindent \textbf{Robot Environment Setup.} 
For both simulation and real robot experiments, we use the Kuka LBR IIWA robot, a 7-DoF torque-controlled robot. Three sensor modalities are available in both simulation and real hardware, including proprioception, an RGB camera, and a force-torque sensor. The proprioceptive feature is the end-effector pose as well as linear and angular velocity. They are computed using forward kinematics. RGB images are recorded from a fixed camera pointed at the robot. Input images to our model are down-sampled to $128\times 128$. On the real robot, we use the Kinect v2 camera. In simulation, we use CHAI3D~\cite{Conti03} for rendering. The force sensor provides a 6-axis feedback on the forces and moments along the x, y , z axes. On the real robot, we mount an OptoForce sensor between the last joint and the peg. In simulation, the contact between the peg and the box is modeled with SAI 2.0~\cite{conti2016framework}, a real-time physics simulator for rigid articulated bodies with high fidelity contact resolution. 

\noindent \textbf{Reward Design.}
We use the following staged reward function to guide the reinforcement learning algorithm through the different sub-tasks, simplifying the challenge of exploration and improving learning efficiency:
%
\[
r(\mathbf{s}) =
  \begin{mcases}[l@{\ }]
    c_r-\frac{c_r}{2}(\tanh{\lambda\|\mathbf{s}\|}+\tanh{\lambda\|\mathbf{s}_{xy}\|}) &\text{(reaching)}\\
    2-c_a\|\mathbf{s}_{xy}\|_2 \,\qquad\; \text{if} \;  \|\mathbf{s}_{xy}\|_2 \leq \epsilon_1 & \text{(alignment)} \\
    4 - 2(\frac{s_z}{h_d-\epsilon_2}) \qquad\,\, \text{if} \;  s_z < 0&\text{(insertion)} \\
    10 \qquad\qquad\qquad\;\text{if} \;  h_d - |s_z| \leq \epsilon_2 &\text{(completion)},
  \end{mcases}
\]
where $\mathbf{s}=(s_x, s_y, s_z)$ and $\mathbf{s}_{xy}=(s_x, s_y)$ use the peg's current position, $\lambda$ is a constant factor to scale the input to the $\tanh$ function. The target peg position is $(0,0,-h_d)$ with $h_d$ as the height of the hole, and $c_r$ and $c_a$ are constant scale factors.

\noindent \textbf{Evaluation Metrics.} We report the quantitative performance of the policies using the sum of rewards achieved in an episode, normalized by the highest attainable reward. We also provide the statistics of the stages of the peg insertion task that each policy can achieve, and report the percentage of evaluation episodes in the following four categories:
\begin{enumerate}[
    topsep=0pt,
    noitemsep,
    leftmargin=*,
    itemindent=12pt]
    \item \emph{completed insertion}: the peg reaches bottom of the hole;
    \item \emph{inserted into hole}: the peg goes into the hole but has not reached the bottom;
    \item \emph{touched the box}: the peg only makes contact with the box; 
    \item \emph{failed}: the peg fails to reach the box.
\end{enumerate}

\noindent \textbf{Implementation Details.} 
To train each representation model, we collect a multimodal dataset of 100k states and generate the self-supervised annotations. We roll out a random policy as well as a heuristic policy while collecting the data, which encourages the peg to make contact with the box. As the policy runs at 20 Hz, it takes 90 to 120 minutes to collect the data. The representation models are trained for 20 epochs on a Titan V GPU before starting policy learning.



\section{Experiments: Results}
We first conduct an ablative study in simulation to investigate the contributions of individual sensory modalities to learning the multimodal representation and manipulation policy. We then apply our full multimodal model to a real robot, and train reinforcement learning policies for the peg insertion tasks from the learned representations with high sample efficiency. Furthermore, we visualize the representations and provide a detailed analysis of robustness with respect to shape and clearance variations.




\subsection{Simulation Experiments}
 Three modalities are encoded and fused by our representation model: RGB images, force readings, and proprioception (see Fig.~\ref{fig:network_architecture}). To investigate the importance of each modality for contact-rich manipulation tasks, we perform an ablative study in simulation, where we learn the multimodal representations with different combinations of modalities. These learned representations are subsequently fed to the TRPO policies to train on a task of inserting a square peg. We randomize the configuration of the box position and the arm's initial position at the beginning of each episode to enhance the robustness and generalization of the model.

We illustrate the training curves of the TRPO agents in \figref{training_curve}. We train all policies with 1.2k episodes, each lasting 500 steps. We evaluate 10 trials with the stochastic policy every 10 training episodes and report the mean and standard deviation of the episode rewards. 
Our \texttt{Full model} corresponds to the multimodal representation model introduced in \secref{representation-learning}, which takes all three modalities as input.
We compare it with three baselines: \texttt{No vision} masks out the visual input to the network, \texttt{No haptics} masks out the haptic input, and \texttt{No vision  No haptics} leaves only proprioceptive input.
From \figref{training_curve} we observe that the absence of either the visual or force modality negatively affects task completion, with \texttt{No vision No haptics} performing the worst. None of the three baselines has reached the same level of performance as the final model. Among these three baselines, we see that the \texttt{No haptics} baseline achieved the highest rewards. We hypothesize that vision locates the box and the hole, which facilitates the first steps of robot reaching and peg alignment, while haptic feedback is uninformative until after contact is made.

The \texttt{Full model} achieves the highest success rate with nearly 80\% completion rate, while all baseline methods have a completion rate below 5\%. It is followed by the \texttt{No haptics} baseline, which relies solely on the visual feedback. We see that it is able to localize the hole and perform insertion half of the time from only the visual inputs; however, few episodes have completed the full insertion. It implies that the haptic feedback plays a more crucial role in determining the actions when the peg is placed in the hole. The remaining two baselines can often reach the box through random exploration, but are unable to exhibit consistent insertion behaviors.


\begin{figure}
    \centering
\begin{subfigure}{.278\linewidth}
    \centering
    \includegraphics[width=1.0\linewidth]{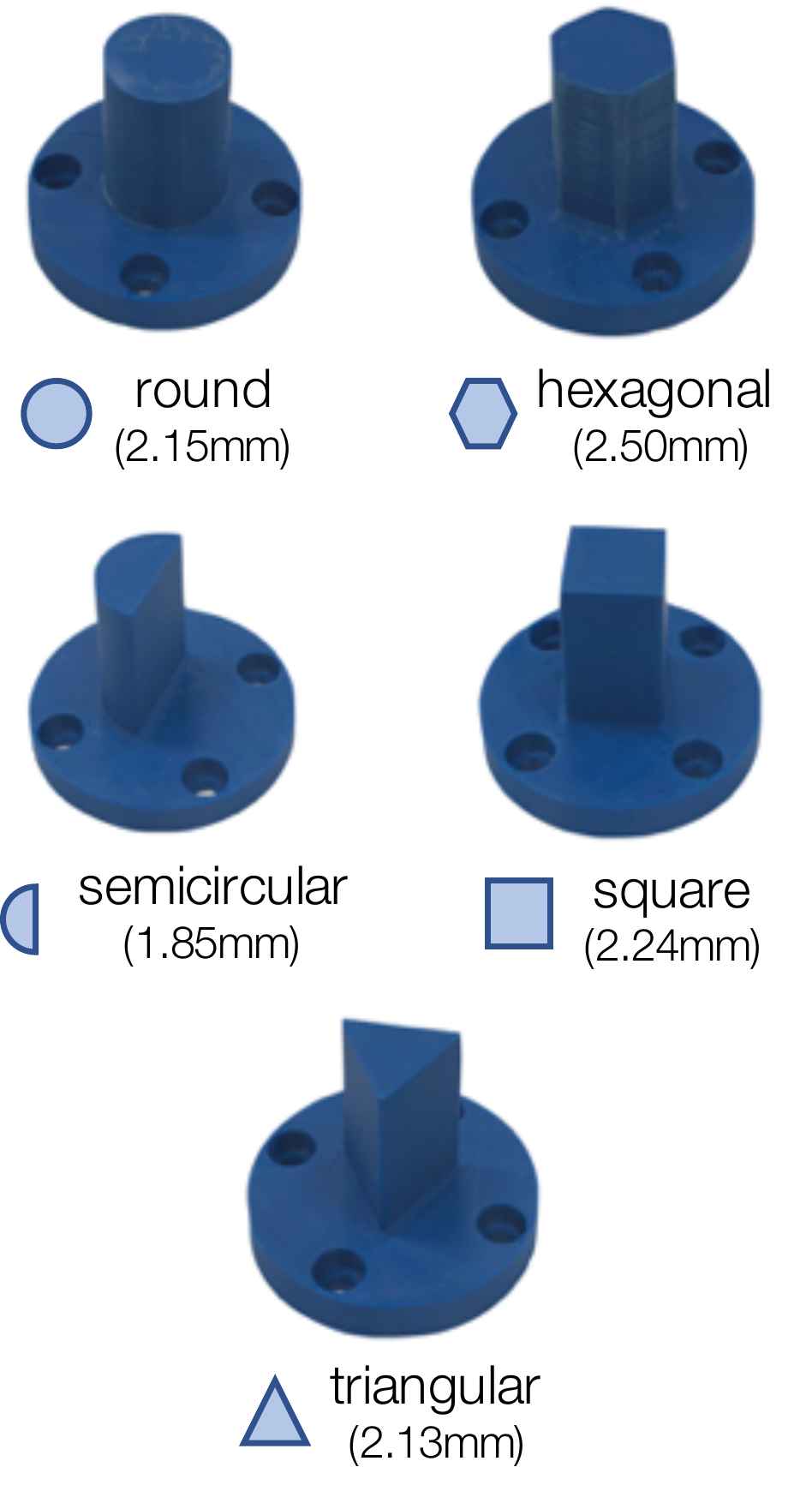}
    \caption{Peg variations}
    \label{fig:peg_types}
\end{subfigure}
\begin{subfigure}{.692\linewidth}
    \centering
    \includegraphics[width=1.0\linewidth]{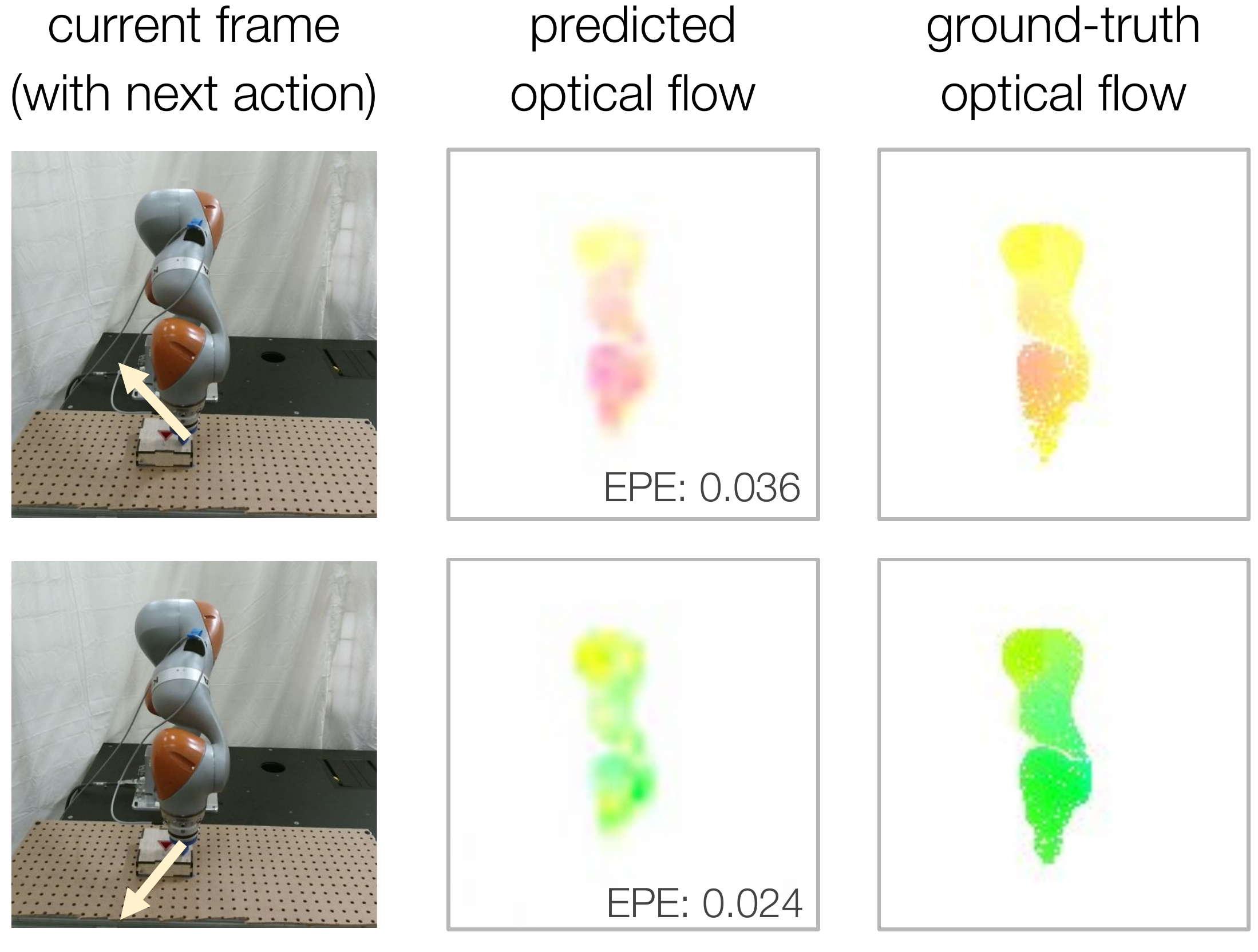}
    \caption{Optical flow prediction examples}
    \label{fig:flow_example}
\end{subfigure}
    
    \caption{(a) 3D printed pegs used in the real robot experiments and their box clearances. (b) Qualitative predictions: We visualize examples of optical flow predictions from our representation model (using color scheme in~\cite{flownet1}). The model predicts different flow maps on the same image conditioned on different next actions indicated by projected arrows.}
    \label{fig:real_robot_info}
\end{figure}

\subsection{Real Robot Experiments}

We evaluate our \texttt{Full model} on the real hardware with round, triangular, and semicircular pegs. In contrast to simulation, the difficulty of sensor synchronization, variable delays from sensing to control, and complex real-world dynamics introduce additional challenges on the real robot. We make the task tractable on a real robot by training a shallow neural network controller while freezing the multimodal representation model that can generate action-conditional flows with low endpoint errors (see \figref{flow_example}).

\begin{figure}
    \centering
    \includegraphics[width=1.0\linewidth]{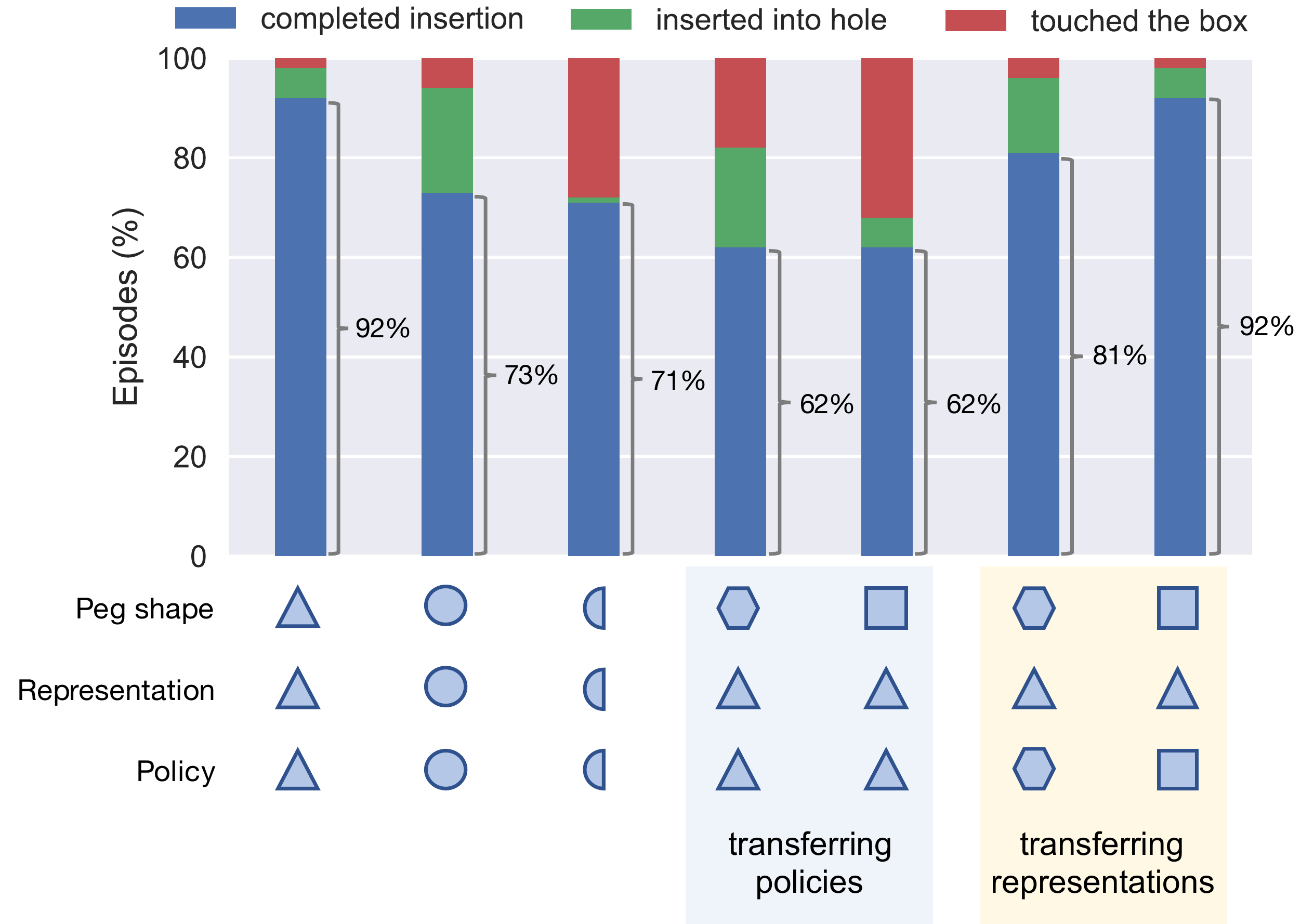}
    \caption{Real Robot Peg Insertion: We evaluate our \texttt{Full Model} on the real hardware with different peg shapes, indicated on the x-axis. The learned policies achieve the tasks with a high success rate. We also study transferring the policies and representations from trained pegs to novel peg shapes (last four bars). The robot effectively re-uses previously trained models to solve new tasks.}
    \label{fig:real_robot_exp}
\end{figure}

We train the TRPO policies for 300 episodes, each lasting 1000 steps, roughly 5 hours of wall-clock time. We evaluate each policy for 100 episodes in \figref{real_robot_exp}. The first three bars correspond to the set of experiments where we train a specific representation model and policy for each type of peg. The robot achieves a level of success similar to that in simulation. A common strategy that the robot learns is to reach the box, search for the hole by sliding over the surface, align the peg with the hole, and finally perform insertion. More qualitative behaviors can be found in the supplementary video.

We further examine the potential of transferring the learned policies and representations to two novel shapes previously unseen in representation and policy training, the hexagonal peg and the square peg. For policy transfer, we take the representation model and the policy trained for the triangular peg, and execute with the new pegs. From the 4th and 5th bars in \figref{real_robot_exp}, we see that the policy achieves over 60\% success rate on both pegs without any further policy training on them. A better transfer performance can be achieved by taking the representation model trained on the triangular peg, and training a new policy for the new pegs. As shown in the last two bars in \figref{real_robot_exp}, the resulting performance increases 19\% for the hexagonal peg and 30\% for the square peg. Our transfer learning results indicate that the multimodal representations from visual and haptic feedback generalize well across variations of our contact-rich manipulation tasks.

Finally, we study the robustness of our policy in the presence of sensory noise and external perturbations to the arm by periodically occluding the camera and pushing the robot arm during trajectory roll-out. The policy is able to recover from both the occlusion and perturbations. Qualitative results can be found in our supplementary video on our website: \url{https://sites.google.com/view/visionandtouch}.

\section{Discussion and Conclusion}


We examined the value of jointly reasoning over time-aligned multisensory data for contact-rich manipulation tasks. To enable efficient real robot training, we proposed a novel model to encode heterogeneous sensory inputs into a compact multimodal representation. Once trained, the representation remained fixed when being used as input to a shallow neural network policy for reinforcement learning. We trained the representation model with self-supervision, eliminating the need for manual annotation. Our experiments with tight clearance peg insertion tasks indicated that they require the multimodal feedback from both vision and touch. We further demonstrated that the multimodal representations transfer well to new task instances of peg insertion. For future work, we plan to extend our method to other contact-rich tasks, which require a full 6-DoF controller of position and orientation. We would also like to explore the value of incorporating richer modalities, such as depth and sound, into our representation learning pipeline, as well as new sources of self-supervision.







\renewcommand*{\bibfont}{\footnotesize}
\begin{flushright}
\printbibliography 
\end{flushright}

\end{document}